\documentclass[11pt,a4paper]{article}
\usepackage[hyperref]{naaclhlt2018}
\usepackage{times}
\usepackage{latexsym}
\usepackage{bm}
\usepackage[pdftex]{graphicx}
\usepackage{amsmath, amssymb}
\usepackage{url}
\usepackage{enumitem}

\aclfinalcopy 
\def\aclpaperid{172} 

\title{Filling Missing Paths: Modeling Co-occurrences of Word Pairs and Dependency Paths for Recognizing Lexical Semantic Relations}

\author{Koki Washio \and Tsuneaki Kato \\
		Department of Language and Information Sciences \\
        Graduate School of Arts and Sciences \\
		The University of Tokyo \\
		3-8-1, Komaba, Meguroku, Tokyo 153-8902 Japan\\
		\tt{\{kokiwashio@g.ecc, kato@boz.c\}.u-tokyo.ac.jp}}

\date{}

\begin{document}
\maketitle
\begin{abstract}
Recognizing lexical semantic relations between word pairs is an important task for many applications of natural language processing.
One of the mainstream approaches to this task is to exploit the lexico-syntactic paths connecting two target words, which reflect the semantic relations of word pairs.
However, this method requires that the considered words co-occur in a sentence.
This requirement is hardly satisfied because of Zipf's law, which states that most content words occur very rarely.
In this paper, we propose novel methods with a neural model of $P(path|w_1, w_2)$ to solve this problem.
Our proposed model of $P(path|w_1, w_2)$ can be learned in an unsupervised manner and can generalize the co-occurrences of word pairs and dependency paths.
This model can be used to augment the path data of word pairs that do not co-occur in the corpus, and extract features capturing relational information from word pairs.
Our experimental results demonstrate that our methods improve on previous neural approaches based on dependency paths and successfully solve the focused problem.
\end{abstract}

\section{Introduction} \label{sec: introduction}
The semantic relations between words are important for many natural language processing tasks, such as recognizing textual entailment \citep{Dagan:2010} and question answering \cite{Yang:2017}.
Moreover, these relations have been also used as features for neural methods in machine translation \citep{sennrich:2016} and relation extraction \citep{Xu:2015}.
This type of information is provided by manually-created semantic taxonomies, such as  WordNet \citep{Fellbaum:1998}.
However, these resources are expensive to expand manually and have limited domain coverage.
Thus, the automatic detection of lexico-semantic relations has been studied for several decades.

One of the most popular approaches is based on patterns that encode a specific kind of relationship (synonym, hypernym, etc.) between adjacent words.
This type of approach is called a path-based method.
Lexico-syntactic patterns between two words provide information on semantic relations.
For example, if we see the pattern, ``animals such as a dog" in a corpus, we can infer that \textit{animal} is a hypernym of \textit{dog}.
On the basis of this assumption, \citet{Hearst:1992} detected the hypernymy relation of two words from a corpus based on several handcrafted lexico-syntactic patterns, e.g., \textit{X such as Y}.
\citet{Snow:2004} used as features indicative dependency paths, in which target word pairs co-occurred, and trained a classifier with data to detect hypernymy relations.

In recent studies, \citet{Shwartz:2016a} proposed a neural path-based model that encoded dependency paths between two words into low-dimensional dense vectors with recurrent neural networks (RNN) for hypernymy detection.
This method can prevent sparse feature space and generalize indicative dependency paths for detecting lexico-semantic relations.
Their model outperformed the previous state-of-the-art path-based method.
Moreover, they demonstrated that these dense path representations capture complementary information with word embeddings that contain individual word features.
This was indicated by the experimental result that showed the combination of path representations and word embeddings improved classification performance.
In addition, \citet{Shwartz:2016b} showed that the neural path-based approach, combined with word embeddings, is effective in recognizing multiple semantic relations.

Although path-based methods can capture the relational information between two words, these methods can obtain clues only for word pairs that co-occur in a corpus.
Even with a very large corpus, it is almost impossible to observe a co-occurrence of arbitrary word pairs.
Thus, path-based methods are still limited in terms of the number of word pairs that are correctly classified.

To address this problem, we propose a novel method with modeling $P(path|w_1, w_2)$ in a neural unsupervised manner, where $w_1$ and $w_2$ are the two target words, and $path$ is a dependency path that can connect the joint co-occurrence of $w_1$ and $w_2$.
A neural model of $P(path|w_1, w_2)$ can generalize co-occurrences of word pairs and dependency paths, and infer plausible dependency paths which connect two words that do not co-occur in a corpus.
After unsupervised learning, this model can be used in two ways:
\begin{itemize}
\item Path data augmentation through predicting dependency paths that are most likely to co-occur with a given word pair.
\item Feature extraction of word pairs, capturing the information of dependency paths as contexts where two words co-occur.
\end{itemize}
While previous supervised path-based methods used only a small portion of a corpus, combining our models makes it possible to use an entire corpus for learning process.

Experimental results for four common datasets of multiple lexico-semantic relations show that our methods improve the classification performance of supervised neural path-based models.

\section{Background}
\subsection{Supervised Lexical Semantic Relation Detection}
Supervised lexical semantic relation detection represents word pairs $(w_1, w_2)$ as feature vectors $\bm{v}$ and trains a classifier with these vectors based on training data.
For word pair representations $\bm{v}$, we can use the distributional information of each word and path information in which two words co-occur.

Several methods exploit word embeddings \citep{Mikolov:2013,Levy:2014,Pennington:2014} as distributional information.
These methods use a combination of each word's embeddings, such as vector concatenation \citep{Baroni:2012,Roller:2016} or vector difference \citep{Roller:2014,Weeds:2014,Vylomova:2016}, as word pair representations.
While these distributional supervised methods do not require co-occurrences of two words in a sentence, \citet{Levy:2015} notes that these methods do not learn the relationships between two words but rather the separate property of each word, i.e., whether or not each word tends to have a target relation.

In contrast, supervised path-based methods can capture relational information between two words.
These methods represent a word pair as the set of lexico-syntactic paths, which connect two target words in a corpus \citep{Snow:2004}.
However, these methods suffer from sparse feature space, as they cannot capture the similarity between indicative lexico-syntactic paths, e.g., \textit{X is a species of Y} and \textit{X is a kind of Y}.

\subsection{Neural Path-based Method}
\label{subsec:neural path-based method}
A neural path-based method can avoid the sparse feature space of the previous path-based methods \citep{Shwartz:2016a,Shwartz:2016b}.
Instead of treating an entire dependency path as a single feature, this model encodes a sequence of edges of a dependency path into a dense vector using a long short-term memory network (LSTM) \citep{Hochreiter:1997}.

A dependency path connecting two words can be extracted from the dependency tree of a sentence.
For example, given the sentence ``A dog is a mammal,'' with \textit{X = dog} and \textit{Y = mammal}, the dependency path connecting the two words is \texttt{X/NOUN/nsubj/> be/VERB/ROOT/- Y/NOUN/attr/<}.
Each edge of a dependency path is composed of a lemma, part of speech (POS), dependency label, and dependency direction.

\citet{Shwartz:2016a} represents each edge as the concatenation of its component embeddings:
\begin{equation}
	\bm{e} = [\bm{v_}{l}; \bm{v}_{pos}; \bm{v}_{dep}; \bm{v}_{dir}]
\end{equation}
where $\bm{v}_{l}, \bm{v}_{pos}, \bm{v}_{dep}$,and $\bm{v}_{dir}$ represent the embedding vectors of the lemma, POS, dependency label, and dependency direction respectively.
This edge vector $\bm{e}$ is an input of the LSTM at each time step.
Here, $\bm{h_t}$, the hidden state at time step $t$, is abstractly computed as:
\begin{equation}
	\bm{h}_{t} = LSTM(\bm{h}_{t-1}, \bm{e}_{t})
\end{equation}
where $LSTM$ computes the current hidden state given the previous hidden state $\bm{h}_{t-1}$ and the current input edge vector $\bm{e}_{t}$ along with the LSTM architecture.
The final hidden state vector $\bm{o}_{p}$ is treated as the representation of the dependency path $p$.

When classifying a word pair $(w_1, w_2)$, the word pair is represented as the average of the dependency path vectors that connect two words in a corpus:
\begin{eqnarray}
	\bm{v}_{(w_1, w_2)} &=& \bm{v}_{paths(w_1, w_2)} \nonumber \\
    &=& \scalebox{0.9}{$\displaystyle \frac{\sum_{p \in paths(w_1, w_2)} f_{p, (w_1, w_2)} \cdot \bm{o}_{p}}{\sum_{p \in paths(w_1, w_2)} f_{p, (w_1, w_2)}}$}
\end{eqnarray}
where $paths(w_1, w_2)$ is the set of dependency paths that connects $w_1$ and $w_2$ in the corpus, and $f_{p, (w_1, w_2)}$ is the frequency of $p$ in $paths(w_1, w_2)$.
The final output of the network is calculated as follows:
\begin{equation}
	\bm{y} = softmax(\bm{W}\bm{v}_{(w_1, w_2)} + \bm{b})
\end{equation}
where $\bm{W} \in \mathbb{R}^{|c| \times d}$ is a linear transformation matrix, $\bm{b} \in \mathbb{R}^{|c|}$ is a bias parameter, $|c|$ is the number of the output class, and $d$ is the size of $\bm{v}_{(w_1, w_2)}$.

This neural path-based model can be combined with distributional methods.
\citet{Shwartz:2016a} concatenated $\bm{v}_{paths(w_1, w_2)}$ to the word embeddings of $w_1$ and $w_2$, redefining $\bm{v}_{(w_1, w_2)}$ as:
\begin{equation}
	\bm{v}_{(w_1, w_2)} = [\bm{v}_{w_1}; \bm{v}_{paths(w_1, w_2)}; \bm{v}_{w_2}]
\end{equation}
where $\bm{v}_{w_1}$ and $\bm{v}_{w_2}$ are word embeddings of $w_1$ and $w_2$, respectively.
This integrated model, named LexNET, exploits both path information and distributional information, and has high generalization performance for lexical semantic relation detection.

\subsection{Missing Path Problem}
All path-based methods, including the neural ones, suffer from data sparseness as they depend on word pair co-occurrences in a corpus.
However, we cannot observe all co-occurrences of semantically related words even with a very large corpus because of Zipf's law, which states that the frequency distribution of words has a long tail; in other words, most words occur very infrequently \citep{hanks:2009}.
In this paper, we refer to this phenomenon as the missing path problem.

This missing path problem leads to the fact that path-based models cannot find any clues for two words that do not co-occur.
Thus, in the neural path-based method, $paths(w_1, w_2)$ for these word pairs is padded with an empty path, like \texttt{UNK-lemma/UNK-POS/UNK-dep/UNK-dir}.
However, this process makes path-based classifiers unable to distinguish between semantically-related pairs with no co-occurrences and those that have no semantic relation.

In an attempt to solve this problem, \citet{Necsulescu:2015} proposed a method that used a graph representation of a corpus. In this graph, words and dependency relations were denoted as nodes and labeled directed edges, respectively.
From this graph representation, paths linking two target words can be extracted through bridging words, even if the two words do not co-occur in the corpus.
They represent word pairs as the sets of paths linking word pairs on the graph and train a support vector machine classifier with training data, thereby improving recall.
However, the authors reported that this method still suffered from data sparseness.

In this paper, we address this missing path problem, which generally restricts path-based methods, by neural modeling $P(path | w_1, w_2)$.

\section{Method} \label{sec:method}

We present a novel method for modeling $P(path | w_1, w_2)$.
The purpose of this method is to address the missing path problem by generalizing the co-occurrences of word pairs and dependency paths.
To model $P(path|w_1, w_2)$, we used the context-prediction approach \citep{Collobert:2008, Mikolov:2013,Levy:2014,Pennington:2014}, which is a widely used method for learning word embeddings.
In our proposed method, word pairs and dependency paths are represented as embeddings that are updated with unsupervised learning through predicting $path$ from $w_1$ and $w_2$ (Section \ref{subsec:unsupervised}).

After the learning process, our model can be used to (1) augment path data by predicting the plausibility of the co-occurrence of two words and a dependency path (Section \ref{subsec:path data augmentation});
and to (2) extract useful features from word pairs, which reflect the information of co-occurring dependency paths (Section \ref{subsec:feature extractor}).

\begin{figure} 
\centering
\includegraphics[width=7.7cm]{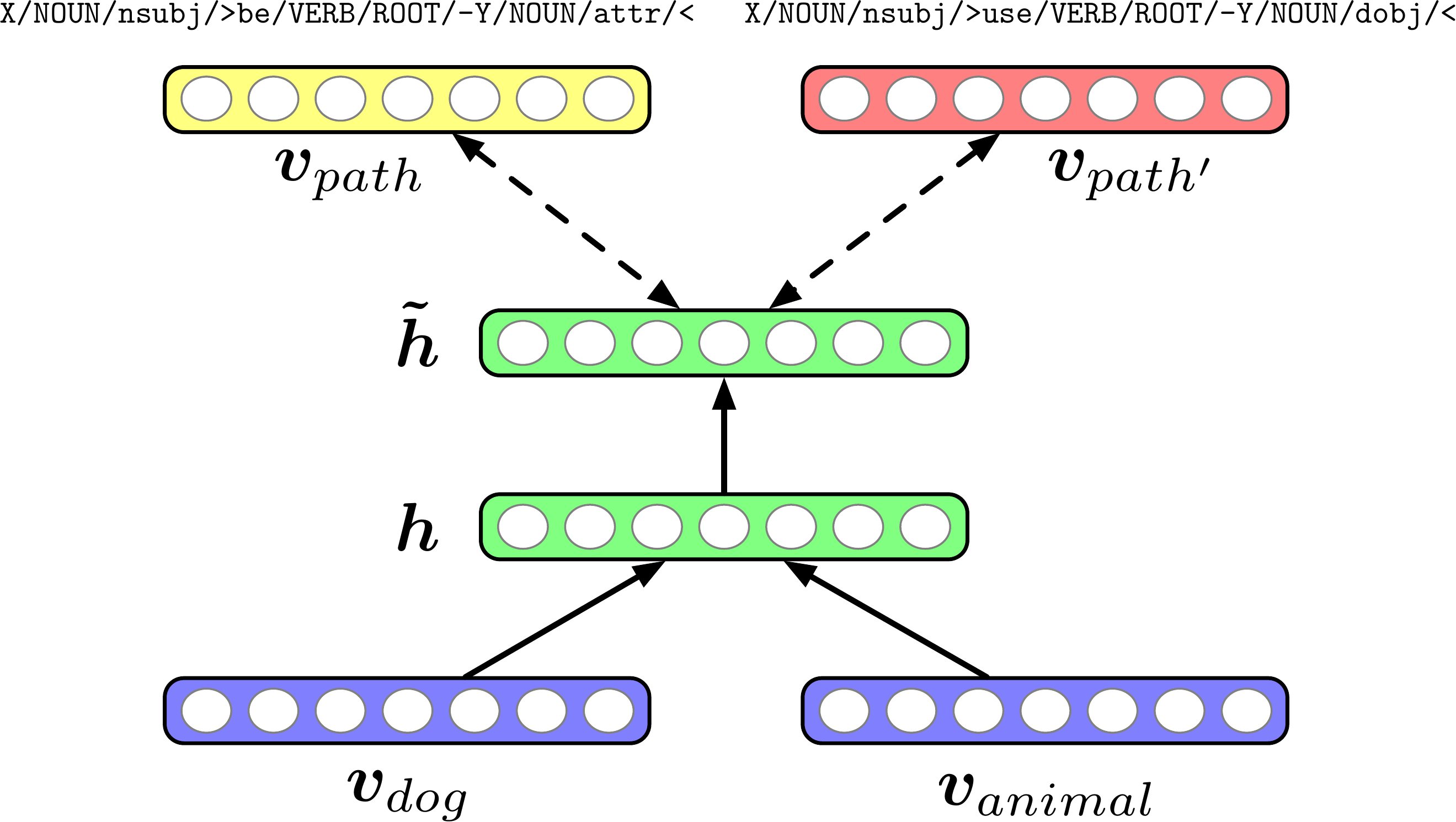}
\caption{
An illustration of our network for modeling $P(path|w_1, w_2)$.
Given a word pair $(dog, animal)$, our model makes $\bm{\tilde{h}}$ of $(dog, animal)$ similar to $\bm{v}_{path}$ of the observed co-occurring dependency path \texttt{X/NOUN/nsubj/> be/VERB/ROOT/- Y/NOUN/attr/<} and dissimilar to $\bm{v}_{path'}$ of the unobserved paths, such as \texttt{X/NOUN/nsubj/> use/VERB/ROOT/- Y/NOUN/dobj/<}, through unsupervised learning.
}
\label{fig:network}
\end{figure}

\subsection{Unsupervised Learning} \label{subsec:unsupervised}

There are many possible ways to model $P(path|w_1, w_2)$.
In this paper, we present a straightforward and efficient architecture, similar to the skip-gram with negative sampling \citep{Mikolov:2013}.

Figure \ref{fig:network} depicts our network structure, which is described below.

\subsubsection*{Data and Network Architecture}

We are able to extract many triples $(w_1, w_2, path)$ from a corpus after dependency parsing.
We denote a set of these triples as $D$.
These triples are the instances used for the unsupervised learning of $P(path|w_1, w_2)$.
Given $(w_1, w_2, path)$, our model learns through predicting $path$ from $w_1$ and $w_2$.

We encode word pairs into dense vectors as follows:
\begin{eqnarray}
	\bm{h}_{(w1,w2)} &=& tanh(\bm{W}_1 [\bm{v}_{w_1}; \bm{v}_{w_2}] + \bm{b}_1) \\
    \bm{\tilde{h}}_{(w_1, w_2)} &=& tanh(\bm{W}_2 \bm{h}_{(w1,w2)} + \bm{b}_2)
\end{eqnarray}
where $[\bm{v}_{w_1}; \bm{v}_{w_2}]$ is the concatenation of the word embeddings of $w_1$ and $w_2$;
$\bm{W}_1$, $\bm{b}_1$, $\bm{W}_2$, and $\bm{b}_2$ are the parameter matrices and bias parameters of the two linear transformations; and
$\bm{\tilde{h}}_{(w_1, w_2)}$ is the representation of the word pair.

We associate each $path$ with the embedding $\bm{v}_{path}$, initialized randomly.
While we use a simple way to represent dependency paths in this paper, LSTM can be used to encode each path in the way described in Section \ref{subsec:neural path-based method}.
If LSTM is used, learning time increases but similarities among paths will be captured.

\subsubsection*{Objective}
We used the negative sampling objective for training \citep{Mikolov:2013}.
Given the word pair representations $\bm{\tilde{h}}_{(w_1, w_2)}$ and the dependency path representations $\bm{v}_{path}$, our model was trained to distinguish real $(w_1, w_2, path)$ triples from incorrect ones.
The log-likelihood objective is as follows:
\footnotesize
\begin{eqnarray}
	L &=& \sum_{(w_1, w_2, path) \in D} \log \sigma(\bm{v}_{path} \cdot \bm{\tilde{h}}_{(w_1, w_2)})  \nonumber \\
    &+& \sum_{(w_1, w_2, path') \in D'} \log \sigma(- \bm{v}_{path'} \cdot \bm{\tilde{h}}_{(w_1, w_2)})
\end{eqnarray}
\normalsize
where, $D'$ is the set of randomly generated negative samples.
We constructed $n$ triples $(w_1, w_2, path')$ for each $(w_1, w_2, path) \in D$, where $n$ is a hyperparameter and each $path'$ is drawn according to its unigram distribution raised to the $3/4$ power.
The objective $L$ was maximized using the stochastic gradient descent algorithm.

\subsection{Path Data Augmentation}
\label{subsec:path data augmentation}
After the unsupervised learning described above, our model of $P(path|w_1, w_2)$ can assign the plausibility score $\sigma(\bm{v}_{path} \cdot \bm{\tilde{h}}_{(w_1, w_2)})$ to the co-occurrences of a word pair and a dependency path.
We can then append the plausible dependency paths to $paths(w_1, w_2)$, the set of dependency paths that connects $w_1$ and $w_2$ in the corpus, based on these scores.

We calculate the score of each dependency path given $(X = w_1, Y = w_2)$ and append the $k$ dependency paths with the highest scores to $paths(w_1, w_2)$, where $k$ is a hyperparameter.
We perform the same process given $(X=w_2, Y=w_1)$ with the exception of swapping the \texttt{X} and \texttt{Y} in the dependency paths to be appended.
As a result, we add $2k$ dependency paths to the set of dependency paths for each word pair.
Through this data augmentation, we can obtain plausible dependency paths even when word pairs do not co-occur in the corpus.
Note that we retain the empty path indicators of $paths(w_1, w_2)$, as we believe that this information contributes to classifying two unrelated words.

\begin{table*}
  \centering
  \begin{tabular}{cc}
    \hline
    \bf{datasets}  & \bf{relations}  \\
    \hline \hline
    K\&H+N  & hypernym, meronym, co-hyponym, random   \\
    BLESS  & hypernym, meronym, co-hyponym, random    \\
    ROOT09  & hypernym, co-hyponym, random  \\
    EVALution  &  hypernym, meronym, attribute, synonym, antonym, holonym, substance meronym  \\
    \hline
  \end{tabular}
  \caption{The relation types in each dataset.}
  \label{table:relation}
\end{table*}

\subsection{Feature Extractor of Word Pairs}
\label{subsec:feature extractor}
Our model can be used as a feature extractor of word pairs.
We can exploit $\bm{\tilde{h}}_{(w_1, w_2)}$ to represent the word pair $(w_1, w_2)$.
This representation captures the information of co-occurrence dependency paths of $(w_1, w_2)$ in a generalized fashion.
Thus, $\bm{\tilde{h}}_{(w_1, w_2)}$ is used to construct the pseudo-path representation $\bm{v}_{p-paths(w_1, w_2)}$.
With our model, we represent the word pair $(w_1, w_2)$ as follows:
\begin{equation}
	\bm{v}_{p-paths(w_1, w_2)} = [\bm{\tilde{h}}_{(w_1, w_2)}; \bm{\tilde{h}}_{(w_2, w_1)}]
\end{equation}
This representation can be used for word pair classification tasks, such as lexical semantic relation detection.

\section{Experiment}
In this section, we examine how our method improves path-based models on several datasets for recognizing lexical semantic relations.
In this paper, we focus on major noun relations, such as hypernymy, co-hypernymy, and meronymy.

\subsection{Dataset}
We relied on the datasets used in \citet{Shwartz:2016b}; K\&H+N \citep{Necsulescu:2015}. BLESS \citep{Baroni:2011}, EVALution \citep{Santus:2015}, and ROOT09 \citep{Santus:2016}.
These datasets were constructed with knowledge resources (e.g., WordNet, Wikipedia), crowd-sourcing, or both.
We used noun pair instances of these datasets.\footnote{We focused only noun pairs to shorten the unsupervised learning time, 
though this restriction is not necessary for our methods and 
the unsupervised learning is still tractable.}
Table \ref{table:relation} displays the relations in each dataset used in our experiments.
Note that we removed the two relations \textit{Entails} and \textit{MemberOf} with few instances from EVALution following \citet{Shwartz:2016b}.
For data splitting, we used the presplitted train/val/test sets from \citet{Shwartz:2016b} after removing all but the noun pairs from each set.

\subsection{Corpus and Dependency Parsing}
\label{subsec:corpus and dependency parsing}

\begin{table}
  \centering
  \small
  \begin{tabular}{|c|c|c|c|}
    \hline
    \bf{datasets} & \bf{instances} & \shortstack{\bf{instances}\\ \bf{with paths}} & \bf{proportion}  \\ \hline
    K\&H+N  & 57509 & 8866 & 15.4\%   \\
    BLESS  & 14558 & 8775 & 60.3\%    \\
    ROOT09  & 8602 & 6582 & 76.5\%  \\
    EVALution  & 3240 & 3199 & 98.7\%  \\
    \hline
  \end{tabular}
  \caption{The number and proportion of instances whose dependency path is obtained from each dataset}
  \label{table:instance}
\end{table}

For path-based methods, we used the June 2017 Wikipedia dump as a corpus and extracted $(w_1, w_2, path)$ triples of noun pairs using the dependency parser of spaCy\footnote{\url{https://spacy.io}} to construct $D$.
In this process, $w_1$ and $w_2$ were lemmatized with spaCy.
We only used the dependency paths which occurred at least five times following the implementation of \citet{Shwartz:2016b}.\footnote{\url{https://github.com/vered1986/LexNET}}

Table \ref{table:instance} displays the number of instances and the proportion of the instances for which at least one dependency path was obtained.

\begin{figure*} 
\centering
\includegraphics[width=16cm]{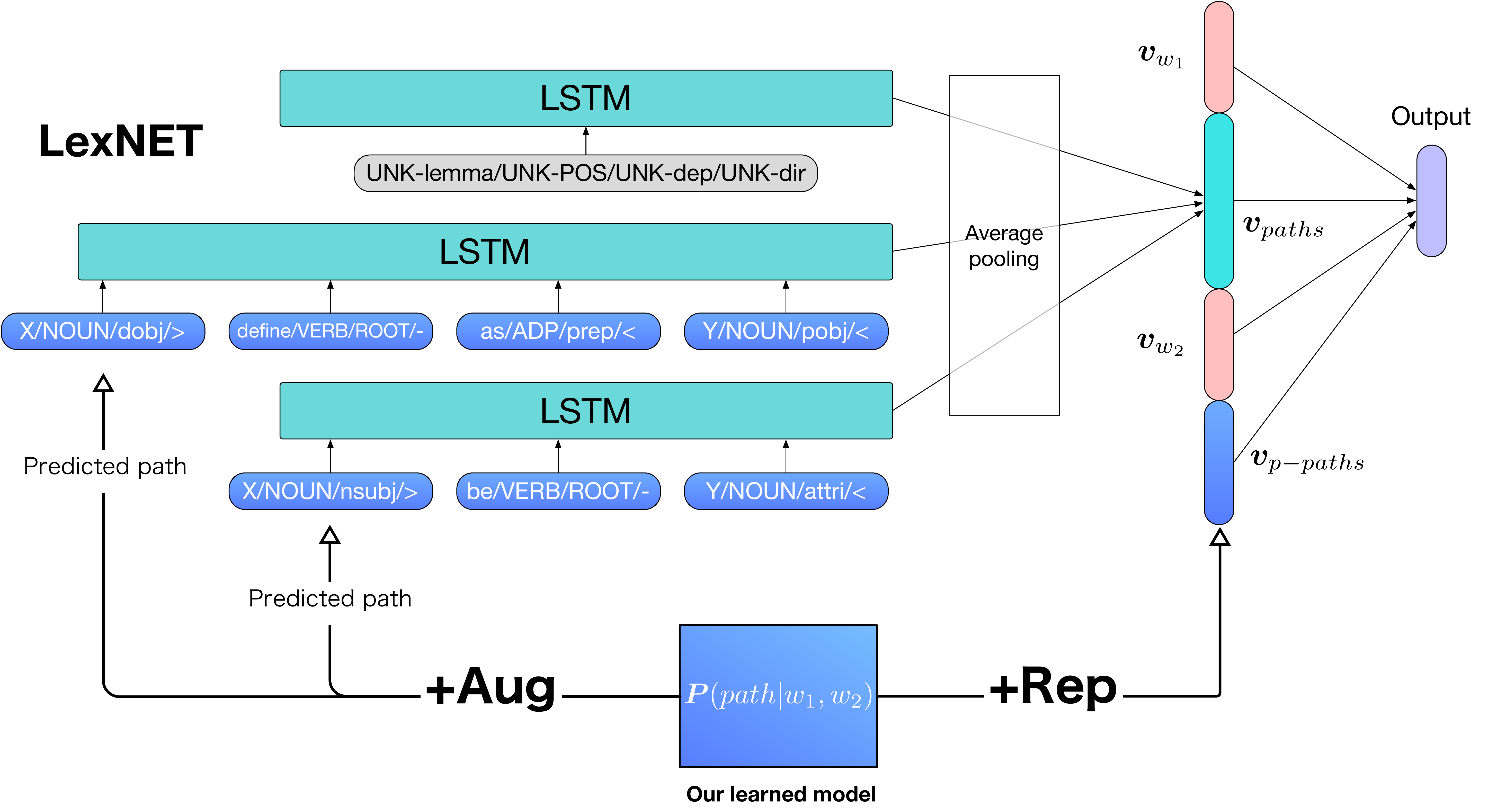}
\caption{
Illustration of +Aug and +Rep applied to LexNET.
+Aug predicts plausible paths from two word embeddings, and these paths are fed into the LSTM path encoder.
+Rep concatenates the pseudo-path representation $\bm{v}_{p-paths(w_1, w_2)}$ with the penultimate layer of LexNET
}
\label{fig:augrep}
\end{figure*}

\subsection{Baseline}

We conducted experiments with three neural path-based methods.
The implementation details below follow those in \citet{Shwartz:2016b}.
We implemented all models using Chainer.\footnote{https://chainer.org}

\begin{description}[style=unboxed,leftmargin=0cm]
\item[Neural Path-Based Model (NPB).]
We implemented and trained the neural path-based model described in Section \ref{subsec:neural path-based method}.
We used the two-layer LSTM with 60-dimensional hidden units.
An input vector was composed of embedding vectors of the lemma (50 dims), POS (4 dims), dependency label (5 dims), and dependency direction (1 dim).
Regularization was applied by a dropout on each of the components’ embeddings \citep{Iyyer:2015,Kiperwasser:2016}.

\item[LexNET.]
We implemented and trained the integrated model LexNET as described in Section \ref{subsec:neural path-based method}. The LSTM details are the same as in the NPB model.

\item[LexNET\_h.]
This model, a variant of LexNET, has an additional hidden layer between the output layer and $\bm{v}_{(w_1, w_2)}$ of Equation (5).
Because of this additional hidden layer, this model can take into account the interaction of the path information and distributional information of two word embeddings.
The size of the additional hidden layer was set to 60. 
\end{description}
Following \citet{Shwartz:2016b}, we optimized each model using Adam (whose learning rate is 0.001) while tuning the dropout rate $dr$ among $\{0.0, 0.2, 0.4\}$ on the validation set.
The minibatch size was set to 100.

We initialized the lemma embeddings of LSTM and concatenated the word embeddings of LexNET with the pretrained 50-dimensional GloVe vector.\footnote{\url{https://nlp.stanford.edu/projects/glove/}}
Training was stopped if performance on the validation set did not improve for seven epochs, and the best model for test evaluation was selected based on the score of the validation set.

\subsection{Our Method}

We implemented and trained our model of $P(path|w_1, w_2)$, described in Section \ref{subsec:unsupervised}, as follows.
We used the most frequent 30,000 paths connecting nouns as the context paths for unsupervised learning.
We initialized word embeddings with the same pretrained GloVe vector as the baseline models.
For unsupervised learning data, we extracted $(w_1, w_2, path)$, whose $w_1$ and $w_2$ are included in the vocabulary of the GloVe vector, and whose $path$ is included in the context paths, from $D$.
The number of these triples was 217,737,765.

We set the size of $\bm{h}_{(w_1, w_2)}$, $\bm{\tilde{h}}_{(w_1,w_2)}$, and $\bm{v}_{path}$ for context paths to 100.
The negative sampling size $n$ was set to 5.
We trained our model for five epochs using Adam (whose learning rate is 0.001).
The minibatch size was 100.
To preserve the distributional regularity of the pretrained word embeddings, we did not update the input word embeddings during the unsupervised learning.

With our trained model, we applied the two methods described in Section \ref{subsec:path data augmentation} and \ref{subsec:feature extractor} to the NPB and LexNET models as follows:
\begin{description}[style=unboxed,leftmargin=0cm]
\item[+Aug.] We added the most plausible $2k$ paths to each $paths(w_1, w_2)$ as in Section \ref{subsec:path data augmentation}. We tuned $k \in \{1, 3, 5\}$ on the validation set.
\item[+Rep.] We concatenated $\bm{v}_{p-paths(w_1, w_2)}$ in Equation (9) with the penultimate layer.
To focus on the pure contribution of unsupervised learning, we did not update this component during supervised learning.
\end{description}
Figure \ref{fig:augrep} illustrates +Aug and +Rep applied to LexNET in the case where the two target words, $w_1$ and $w_2$, do not co-occur in the corpus.

\begin{table*}
  \centering
  \begin{tabular}{|l|cccc|}
    \hline
    \multicolumn{1}{|c|}{\textbf{Models}} & \textbf{K\&H+N} & \textbf{BLESS} & \textbf{ROOT09} & \textbf{EVALution}  \\ \hline
    NPB & 0.495 & 0.773 & 0.731 & 0.463 \\
    NPB+Aug & \textbf{0.897} & \textbf{0.842} & \textbf{0.778} & \textbf{0.489} \\
    \hline
  \end{tabular}
  \caption{Classification performance of the neural path-based model (NPB) and that with the path data augmentation (NPB+Aug).}
  \label{table:path-based aug}
\end{table*}

\begin{table*}
  \centering
  \begin{tabular}{|l|cccc|}
    \hline
    \multicolumn{1}{|c|}{\textbf{Models}} & \textbf{K\&H+N} & \textbf{BLESS} & \textbf{ROOT09} & \textbf{EVALution}  \\ \hline
    LexNET & 0.969 & 0.922 & 0.776 & 0.539 \\
    LexNET\_h & 0.968 & 0.927 & 0.810 & 0.540 \\
    \hline
    LexNET+Aug & \textbf{0.970} & 0.927 & 0.806 & 0.545 \\
    LexNET+Rep & \textbf{0.970} & \textbf{0.944} & \textbf{0.832} & 0.565 \\
    LexNET+Aug+Rep & 0.969 & 0.942 & 0.820 & \textbf{0.567} \\ 
    \hline
  \end{tabular}
  \caption{Classification performance of the integrated model, LexNET and LexNET\_h, and those with our methods, +Aug and +Rep.}
  \label{table:integrated aug rep}
\end{table*}

\section{Result}
In this section we examine how our methods improved the baseline models.
Following the previous research \citep{Shwartz:2016b}, the performance metrics were the ``averaged'' $F1$ of scikit-learn \citep{Pedregosa:2011}, which computes the $F1$ for each relation, and reports their average weighted by the number of true instances for each relation.

\subsection{Path-based Model and Path Data Augmentation}
We examined whether or not our path data augmentation method +Aug contributes to the neural path-based method.
The results are displayed in Table \ref{table:path-based aug}.

Applying our path data augmentation method improved the classification performance on each dataset.
Especially for K\&H+N, the large dataset where the three-fourths of word pairs had no paths, our method significantly improved the performance. 
This result shows that our path data augmentation effectively solves the missing path problem.
Moreover, the model with our method outperforms the baseline on EVALution, in which nearly all word pairs co-occurred in the corpus.
This indicates that the predicted paths provide useful information and enhance the path-based classification.
We examine the paths that were predicted by our model of $P(path|w_1, w_2)$ in Section \ref{subsec:predicted dependency paths}.

\subsection{Integrated Model and Our Methods}

We investigated how our methods using modeling $P(path|w_1, w_2)$ improved the baseline integrated model, LexNET.
Table \ref{table:integrated aug rep} displays the results.

Our proposed methods, +Aug and +Rep, improved the performance of LexNET on each dataset.\footnote{The improvement for K\&H+N is smaller than those for
the others. We think this owes to most instances
of this dataset being correctly classified only by distributional
information. This view is supported by \citet{Shwartz:2016b},
in which LexNET hardly outperformed a distributional
method for this dataset.}
Moreover, the best score on each dataset was achieved by the model to which our methods were applied.
These results show that our methods are also effective with the integrated models based on path information and distributional information.

The table also shows that LexNET+Rep outperforms LexNET\_h, though the former has fewer parameters to be tuned during the supervised learning than the latter. This indicates that the word pair representations of our model capture information beyond the interaction of two word embeddings.
We investigate the properties of our word pair representation in Section \ref{subsec:visualizing}.

Finally, We found that applying both methods did not necessarily yield the best performance.
A possible explanation for this is that applying both methods is redundant, as both +Aug and +Rep depend on the same model of $P(path|w_1, w_2)$.

\section{Analysis}

\begin{table*}
  \centering
  \scalebox{0.9}{
  \small
  \begin{tabular}{|c|c|p{9.5cm}|}
    \hline
    \multicolumn{1}{|c|}{\textbf{Word pair}} & \textbf{Relation} & \multicolumn{1}{|c|}{\textbf{Predicted paths}}   \\ \hline
    & & \textbf{\texttt{X/NOUN/nsubj/> be/VERB/ROOT/- shooter/NOUN/attr/< Y/NOUN/compound/<}} \\ \cline{3-3}
    X = ``jacket'', Y = ``commodity'' & hypernym & \textbf{\texttt{X/NOUN/nsubj/> be/VERB/ROOT/- Y/NOUN/attr/< manufacture/VERB/acl/<}} \\ \cline{3-3}
    & & \textbf{\texttt{red/ADJ/amod/< X/NOUN/nsubj/> be/VERB/ROOT/- Y/NOUN/attr/<}} \\ \hline
    
    & & \textbf{\texttt{X/NOUN/nsubj/> be/VERB/ROOT/- species/NOUN/attr/< of/ADP/prep/< Y/NOUN/pobj/< of/ADP/prep/>}} \\ \cline{3-3}
    X = ``goose'', Y = ``creature'' & hypernym & \textbf{\texttt{X/NOUN/nsubj/> be/VERB/ROOT/- specie/NOUN/attr/< of/ADP/prep/< Y/NOUN/pobj/< in/ADP/prep/>}} \\ \cline{3-3}
    & & \texttt{X/NOUN/pobj/> of/ADP/ROOT/- bird/NOUN/pobj/< Y/NOUN/conj/<} \\ \hline
    \hline
    
    & & \textbf{\texttt{X/NOUN/ROOT/- represent/VERB/relcl/< Y/NOUN/nsubj/<}} \\ \cline{3-3}
    X = ``owl'', Y = ``rump'' & meronym & \textbf{\texttt{X/NOUN/nsubj/> have/VERB/ROOT/- Y/NOUN/dobj/< be/VERB/relcl/>}} \\ \cline{3-3}
    & & \textbf{\texttt{all/DET/det/< X/NOUN/nsubj/> have/VERB/ROOT/- Y/NOUN/dobj/<}} \\ \hline
    
    & & \texttt{X/NOUN/pobj/> of/ADP/ROOT/- arm/NOUN/pobj/< Y/NOUN/conj/<} \\ \cline{3-3}
    X = ``mug'', Y = ``plastic'' & meronym & \textbf{\texttt{the/DET/det/< X/NOUN/nsubjpass/> make/VERB/ROOT/- from/ADP/prep/< Y/NOUN/pobj/<}} \\ \cline{3-3}
    & & \texttt{X/NOUN/compound/> gun/NOUN/ROOT/- Y/NOUN/appos/<} \\ \hline
    \hline
    
    & & \textbf{\texttt{X/NOUN/compound/> leaf/NOUN/ROOT/- Y/NOUN/conj/<}} \\ \cline{3-3}
    X = ``carrot'', Y = ``beans'' & co-hyponym & \textbf{\texttt{X/NOUN/compound/> specie/NOUN/ROOT/- Y/NOUN/conj/<}} \\ \cline{3-3}
    & & \texttt{X/NOUN/dobj/> use/VERB/ROOT/- in/ADP/prep/< Y/NOUN/pobj/< of/ADP/prep/>} \\ \hline
    
    & & \textbf{\texttt{X/NOUN/dobj/> play/VERB/ROOT/- guitar/NOUN/dobj/< Y/NOUN/conj/<}} \\ \cline{3-3}
    X = ``cello'', Y = ``kazoo'' & co-hyponym & \textbf{\texttt{X/NOUN/pobj/> for/ADP/ROOT/- piano/NOUN/pobj/< Y/NOUN/conj<}} \\ \cline{3-3}
    & & \textbf{\texttt{X/NOUN/pobj/> on/ADP/ROOT/- drum/NOUN/pobj/< Y/NOUN/conj/<}} \\ 
    
   \hline
  \end{tabular}
  }
  \caption{Predicted paths with our model for a word pair of each relation in BLESS.}
  \label{table:predicted paths}
\end{table*}

\label{sec:analysis}
In this section, we investigate the properties of the predicted dependency paths and word pair representations of our model.

\subsection{Predicted Dependency Paths}
\label{subsec:predicted dependency paths}
We extracted the word pairs of BLESS without co-occurring dependency paths and predicted the plausible dependency paths of those pairs with our model of $P(path|w_1, w_2)$.
The examples are displayed in Table \ref{table:predicted paths} at the top three paths.
We used the bold style for the paths that we believe to be indicative or representative for a given relationship.

Our model predicted plausible and indicative dependency paths for each relation, although the predicted paths also contain some implausible or unindicative ones.
For hypernymy, our model predicted variants of the is-a path according to domains, such as \textit{X is Y manufactured} in the clothing domain and \textit{X is a species of Y} in the animal domain.
For $(owl, rump)$, which is a meronymy pair, the top predicted path was \textit{X that Y represent}.
This is not plausible for $(owl, rump)$ but is indicative for meronymy, particularly member-of relations.
Moreover, domain-independent paths which indicate meronymy, such as \textit{all X have Y}, were predicted.
For $(mug, plastic)$, one of the predicted paths, \textit{X is made from Y}, is also a domain-independent indicative path for meronymy.
For co-hypernymy, our model predicted domain-specific paths, which indicate that two nouns are of the same kind.
For examples, given \textit{X leaf and Y} and \textit{X specie and Y} of $(carrot, beans)$, we can infer that both X and Y are plants or vegetables.
Likewise, given \textit{play X, guitar, and Y} of $(cello, kazoo)$, we can infer that both X and Y are musical instruments.
These examples show that our path data augmentation is effective for the missing path problem and enhances path-based models.

\subsection{Visualizing Word Pair Representations}
\label{subsec:visualizing}

\begin{figure*} 
\centering
\includegraphics[width=16cm]{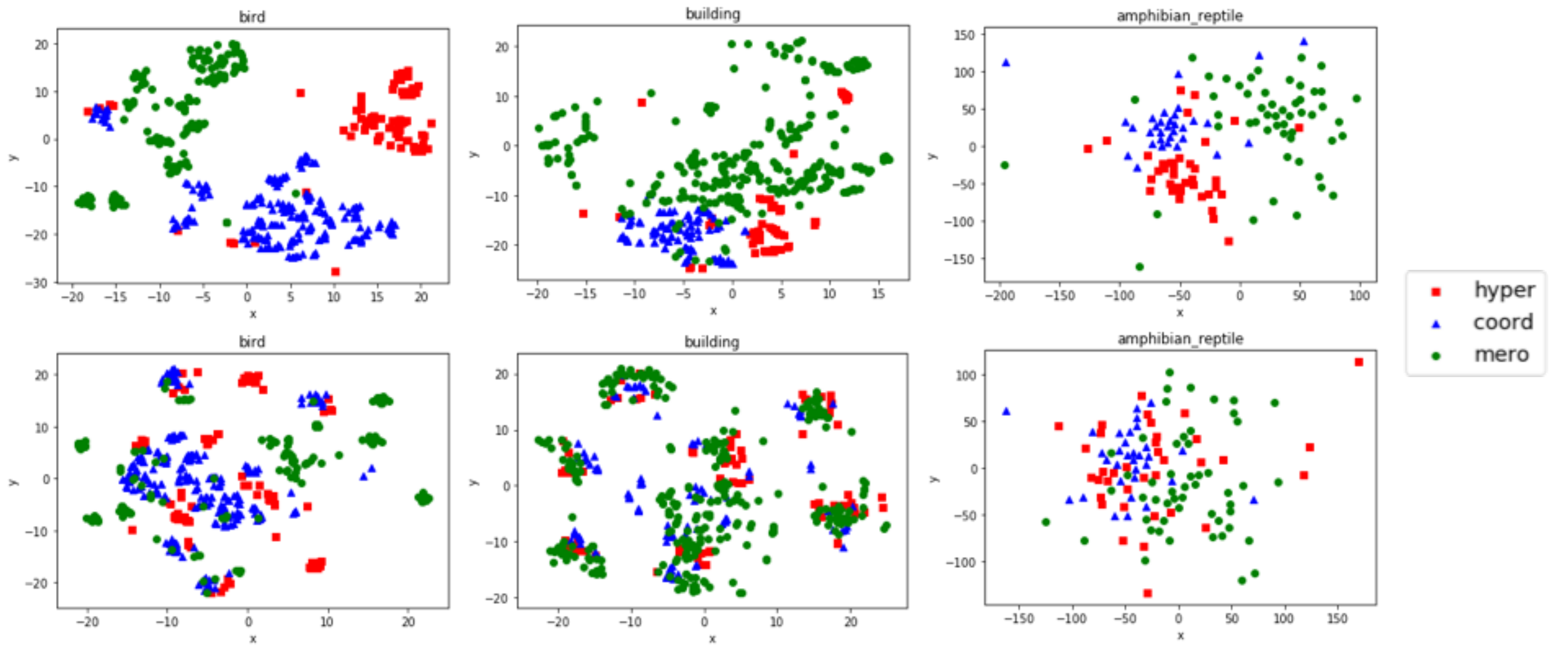}
\caption{
Visualization of the our word pair representations $\bm{v}_{p-paths(w_1, w_2)}$ (top row) and the concatenation of two word embeddings (bottom row) using t-SNE in some domains.
The two axes of each plot, $x$ and $y$, are the reduced dimensions using t-SNE.}
\label{fig:tsne}
\end{figure*}

We visualized the word pair representations $\bm{v}_{p-paths(w_1, w_2)}$ to examine their specific properties.
In BLESS, every pair was annotated with 17 domain class labels.
For each domain, we reduced the dimensionality of the representations using t-SNE \citep{Maaten:2008} and plotted the data points of the hypernyms, co-hyponyms, and meronyms.
We compared our representations with the concatenation of two word embeddings (pretrained 50-dimensional GloVe).
The examples are displayed in Figure \ref{fig:tsne}.

We found that our representations (the top row in Figure \ref{fig:tsne}) grouped the word pairs according to their semantic relation in some specific domains based only on unsupervised learning.
This property is desirable for the lexical semantic relation detection task.
In contrast to our representations, the concatenation of word embeddings (the bottom row in Figure \ref{fig:tsne}) has little or no such tendency in all domains.
The data points of the concatenation of word embeddings are scattered or jumbled.
This is because the concatenation of word embeddings cannot capture the relational information of word pairs but only the distributional information of each word \citep{Levy:2015}.

This visualization further shows that our word pair representations can be used as pseudo-path representations to alleviate the missing path problem.

\section{Conclusion}
In this paper, we proposed the novel methods with modeling $P(path|w_1, w_2)$ to solve the missing path problem.
Our neural model of $P(path|w_1, w_2)$ can be learned from a corpus in an unsupervised manner, and can generalize co-occurrences of word pairs and dependency paths.
We demonstrated that this model can be applied in the two ways: (1) to augment path data by predicting plausible paths for a given word pair, and (2) to extract from word pairs useful features capturing co-occurring path information.
Finally, our experiments demonstrated that our methods can improve upon the previous models and successfully solve the missing path problem.

In future work, we will explore unsupervised learning with a neural path encoder.
Our model bears not only word pair representations but also dependency path representations as context vectors.
Thus, we intend to apply these representations to various tasks, which path representations contribute to.

\section*{Acknowledgments}

This work was supported by JSPS KAKENHI Grant numbers JP17H01831, JP15K12873.




\bibliography{naaclhlt2018}
\bibliographystyle{acl_natbib}

\end{document}